\documentclass[times,twocolumn,final]{elsarticle}
\usepackage{ipcai_arxiv}
\usepackage{framed,multirow}

\usepackage{amssymb}
\usepackage{latexsym}

\usepackage{url}
\usepackage{cite}
\usepackage{amsmath,amssymb,amsfonts,bm}
\usepackage{booktabs}
\usepackage{arydshln}
\usepackage{tabulary}
\usepackage{graphicx,epstopdf}
\usepackage{textcomp}
\usepackage{subcaption}
\usepackage{booktabs}
\usepackage[svgnames,table]{xcolor}
\usepackage{pifont}
\usepackage{arydshln}
\usepackage[colorlinks=true,citecolor=cyan,urlcolor=cyan]{hyperref}
\usepackage{graphicx}
\usepackage{placeins}
\usepackage{fancyhdr}
\usepackage{float}

\usepackage{listings}%
\usepackage{lipsum}
\usepackage{makecell}

\fancypagestyle{firstpagestyle}{
    \fancyhf{} % Clear header and footer
    \fancyhead{} % Clear header
    \fancyfoot{} % Clear footer
    \fancyhead[CO]{\em \fontsize{9pt}{8pt}\selectfont This article has been accepted for publication in IEEE Transactions on Medical Imaging (TMI).}
}

\begin{document}

% \begin{frontmatter}

\title{Gaze Graphs: A Scene-Level Representation of Visual Attention in the Operating Room}

\author[1,2]{Séraphin \snm{Baributsa}}

\author[1,2]{Keqi \snm{Chen}\corref{cor}}

\author[1,2,3]{Vinkle \snm{Srivastav}}

\author[1,2]{Nicolas \snm{Padoy}}

\cortext[cor]{Corresponding author: keqi.chen@unistra.fr}

\address[1]{University of Strasbourg, CNRS, INSERM, ICube, UMR7357, France}

\address[2]{IHU Strasbourg, 67000 Strasbourg, France}

\address[3]{Indian Institute of Technology (IIT) Madras, 600036 Chennai, India}

\received{XXX}
\finalform{XXX}
\accepted{XXX}
\availableonline{XXX}
\communicated{XXX}

\begin{abstract}

\textbf{Purpose: } 
Analyzing clinicians' visual attention from exocentric camera views can support surgical scene understanding without requiring wearable eye trackers. However, existing gaze-following research in the operating room (OR) estimates only 2D gaze locations without identifying attended scene entities or localizing gaze targets in 3D space. In this work, we investigate gaze graphs as a scene-level representation of visual attention that connects clinicians to attended scene entities, and evaluate their utility for fine-grained surgical action recognition and semantic scene graph (SSG) generation.

\textbf{Methods: } 
We extend the public EgoExOR dataset with gaze annotations to create \emph{EgoExOR-Gaze}, the first benchmark for multi-view gaze graph generation in the OR.  We propose a novel architecture that jointly estimates 3D gaze targets and predicts attended entities using only exocentric camera views. Specifically, our approach exploits 3D gaze geometry to sample local visual features across multiple views for attended entity prediction. We further utilize predicted gaze graphs for fine-grained surgical action recognition and SSG generation. 

\textbf{Results: } 
For gaze graph generation, our model outperforms both exocentric-only and combined exocentric-egocentric baselines. For surgical action recognition, our model incorporating predicted gaze graphs outperforms both exocentric and egocentric baselines across all evaluated metrics, with particularly strong gains in temporal metrics. For SSG generation, incorporating gaze graphs improves macro-F1 from 0.42 to 0.52 compared to exocentric inputs alone, although performance remains below that achieved with egocentric inputs.

\textbf{Conclusion: } 
Our results demonstrate the value of gaze graphs for surgical action recognition and SSG generation, highlighting the importance of explicitly modeling scene-level visual attention even when egocentric views are unavailable.
\end{abstract}
%\end{frontmatter}

\maketitle

\begin{figure*}
\centering
\includegraphics[trim=0 85 0 85, clip, width=0.8\linewidth]{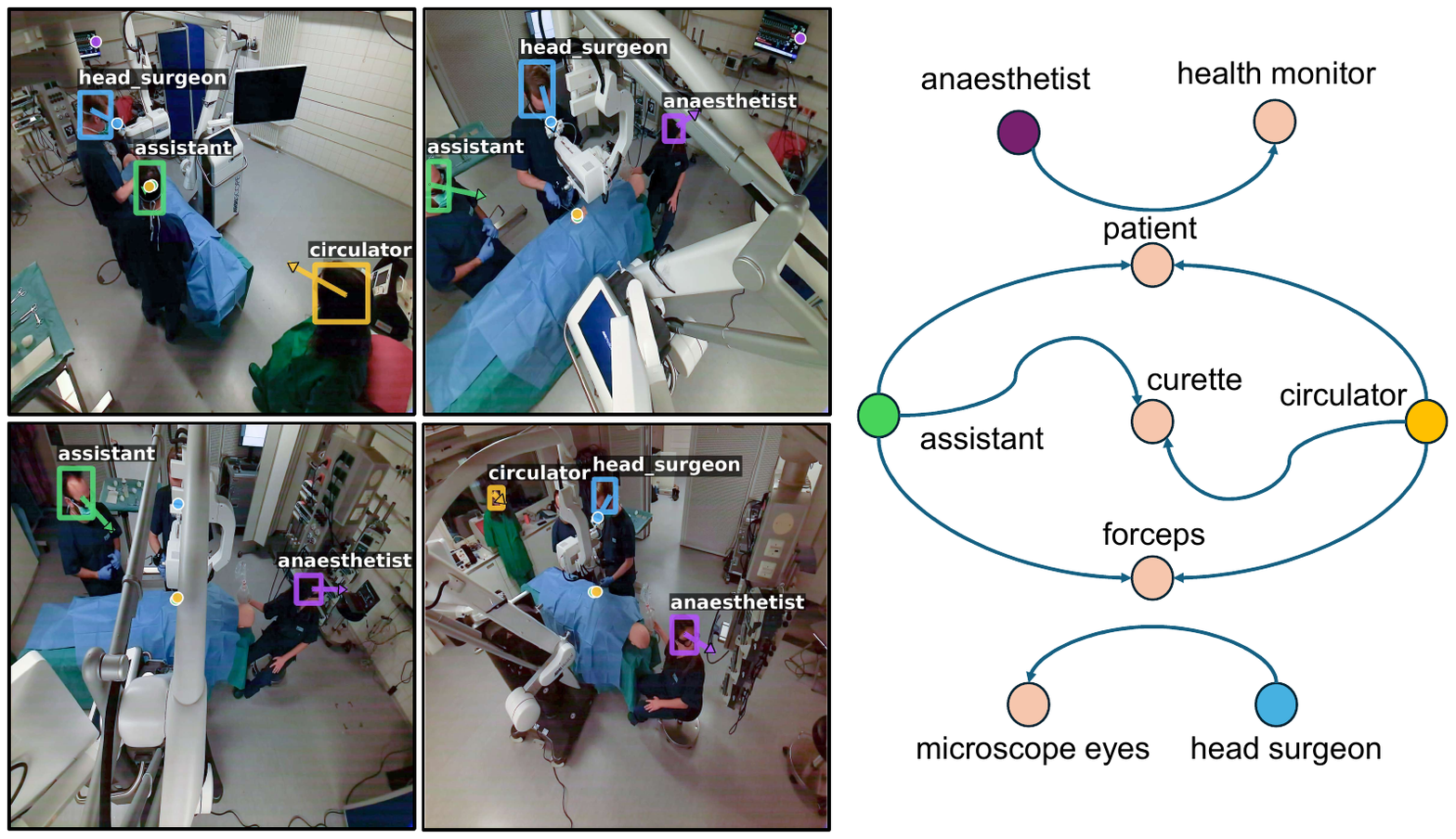}
\caption{An example of dataset annotations. On the left, we show the annotated head bounding box, clinical role, and corresponding gaze point of each person. On the right, we show the annotated gaze graph where arrows denote ``looking'' relations pointing from clinical roles to one or multiple entities within the attended regions.}
\label{fig:annotations}
\end{figure*}

\section{Introduction}\label{sec:intro}

The operating room (OR) is a highly dynamic, collaborative environment where multiple clinicians interact with patients, surgical instruments, medical equipment, and one another throughout surgical procedures. Understanding the relationships among these scene entities is fundamental to interpreting surgical activities and workflows, ultimately supporting the development of context-aware assistance systems~\citep{ozsoy2025mm}. Existing approaches commonly characterize these relationships through observable actions and spatial configurations, such as holding surgical instruments, operating on patients, or standing near the operating table~\citep{ozsoy20224d}. However, these relations do not explicitly represent clinicians' visual attention, which plays an important role in situation awareness in the OR~\citep{tien2010measuring,fioratou2010beyond}. For instance, clinicians may observe their colleagues' actions without participating, monitor medical equipment without physical interaction, or share a common visual focus while performing different tasks. These examples motivate scene-level gaze analysis, which captures attentional relationships and provides complementary information for understanding surgical scenes. 

Over the past decades, researchers have investigated clinicians' visual attention using wearable or head-mounted eye-tracking systems~\citep{seagull1999monitoring,atkins2013surgeons,erridge2018comparison,ozsoy2026egoexor}. However, collecting synchronized egocentric gaze data from the OR team requires configuring and coordinating multiple eye trackers, making large-scale deployment technically challenging. Moreover, wearing these devices during surgical procedures may disrupt clinical workflows and pose challenges to maintaining sterile conditions. To avoid eye trackers, researchers have introduced gaze-following in the surgical domain to predict where clinicians are looking using exocentric camera views~\citep{chen2026they}. These gaze predictions have demonstrated their value in downstream tasks, including clinical role prediction, surgical phase recognition, and team communication detection. Nevertheless, this work estimates gaze targets only as 2D locations in individual camera views, without explicitly identifying the attended scene entities or recovering their 3D positions. These limitations make it difficult to represent clinicians' visual attention toward scene entities consistently across multiple camera views.

\begin{table}[t]
\centering
\caption{Summary of the gaze graph dataset splits and annotations.}
\label{tab:dataset_stats}
\resizebox{0.48\textwidth}{!}{
\begin{tabular}{l|ccccc}
\toprule
Split & Takes & Frames & Head boxes & 3D gaze points & \emph{looking} relations \\
\midrule
Train      & 19 & 2,829 & 26,664 & 6,568 & 12,057 \\
Validation & 6  & 771   & 7,651  & 1,979& 3,614  \\
Test       & 5  & 832   & 7,825  & 1,935& 3,517  \\
\midrule
Total      & 30 & 4,432 & 42,140 & 10,482& 19,188 \\
\bottomrule
\end{tabular}
}
\end{table}

Beyond predicting 2D gaze points, gaze object detection has been explored in general computer vision and aims to identify the objects individuals are looking at~\citep{wang2022gatector}. However, these approaches primarily focus on individual gaze targets rather than modeling attentional relations at the scene level. For scene-level understanding, researchers have explored gaze-based graph representations for social interaction recognition~\citep{malik2023relational}, and introduced semantic scene graphs (SSGs) into the OR to represent scene entities and their relations~\citep{ozsoy20224d,ozsoy2026egoexor}. Inspired by these works, we investigate gaze graphs as a specialized form of SSG that focuses exclusively on ``looking'' relations between clinicians and scene entities, rather than the broader range of relations modeled by existing surgical SSGs. We hypothesize that gaze graphs can serve not only as an effective representation of scene-level visual attention for downstream surgical workflow analysis, but also as an intermediate representation for generating more comprehensive SSGs.

To investigate these hypotheses, we build upon the EgoExOR dataset~\citep{ozsoy2026egoexor} by extending its annotations with complete ``looking'' relations, head bounding boxes, and 3D gaze target points. We then propose a novel multi-view architecture for gaze graph generation using only exocentric views. Our approach outperforms existing SSG generation baselines in predicting ``looking'' relations. We further evaluate the generated gaze graphs on two downstream tasks. First, we incorporate gaze graphs as additional inputs to a transformer-based model for fine-grained surgical action recognition on the OR-Action benchmark~\citep{tristram2026or}. Second, we explore the use of gaze graphs as a practical alternative to egocentric views for SSG generation. Experimental results show that generated gaze graphs substantially improve fine-grained surgical action recognition. For SSG generation, incorporating gaze graphs improves performance over using exocentric views alone, but does not fully match the performance achieved with egocentric views, particularly for relations requiring fine-grained first-person visual information. 

Our contributions can be summarized as follows: (1) we establish the first benchmark for multi-view gaze graph generation in the OR by extending EgoExOR with comprehensive gaze annotations; (2) we present a novel transformer-based multi-view approach for gaze graph generation using only exocentric views, which outperforms existing SSG generation baselines in predicting ``looking'' relations; (3) we demonstrate the effectiveness of generated gaze graphs for fine-grained surgical action recognition on the OR-Action benchmark, achieving state-of-the-art performance; (4) we investigate the use of generated gaze graphs as an alternative to egocentric visual inputs for SSG generation and analyze performance differences across relation types.

\begin{table*}[t]
\centering
\caption{Number of ``looking'' relations per target entity class and split.}
\label{tab:entity_dist}
\resizebox{0.8\textwidth}{!}{
\begin{tabular}{@{}lrrrr@{\hspace{8pt}}lrrrr@{}}
\toprule
Entity & Train & Val & Test & Total &
Entity & Train & Val & Test & Total \\
\midrule
patient
& 3,010 & 946 & 976 & 4,932
& anesthesia\_equipment
& 121 & 34 & 49 & 204 \\

ultrasound\_machine
& 1,735 & 381 & 415 & 2,531
& dressing\_material
& 58 & 130 & 0 & 188 \\

ultrasound\_probe
& 1,592 & 390 & 520 & 2,502
& assistant
& 100 & 31 & 44 & 175 \\

ultrasound\_screen
& 1,583 & 352 & 389 & 2,324
& head\_surgeon
& 91 & 42 & 29 & 162 \\

needle
& 883 & 263 & 300 & 1,446
& curette
& 91 & 65 & 0 & 156 \\

syringe
& 564 & 180 & 178 & 922
& scalpel
& 56 & 33 & 31 & 120 \\

instrument\_table
& 364 & 135 & 111 & 610
& microscope\_controller
& 59 & 30 & 22 & 111 \\

microscope
& 252 & 104 & 55 & 411
& body\_marker
& 63 & 0 & 22 & 85 \\

forceps
& 200 & 107 & 49 & 356
& operating\_table
& 42 & 1 & 11 & 54 \\

health\_monitor
& 200 & 67 & 72 & 339
& bin
& 30 & 15 & 4 & 49 \\

tissue\_paper
& 214 & 52 & 51 & 317
& anaesthetist
& 21 & 7 & 20 & 48 \\

ultrasound\_gel
& 212 & 45 & 51 & 308
& antiseptic
& 11 & 0 & 23 & 34 \\

cotton
& 149 & 65 & 50 & 264
& circulator
& 23 & 4 & 3 & 30 \\

microscope\_screen
& 140 & 63 & 37 & 240
& unsterile\_instrument
& 19 & 0 & 0 & 19 \\

microscope\_eye
& 161 & 72 & 5 & 238
& tissue\_mark
& 13 & 0 & 0 & 13 \\
\midrule
\multicolumn{6}{l}{\textbf{All entities}}
& 12,057 & 3,614 & 3,517 & 19,188 \\
\bottomrule
\end{tabular}
}
\end{table*}

\section{EgoExOR-Gaze dataset}\label{sec:dataset}

To enable gaze graph study in the OR, we extend the EgoExOR dataset~\citep{ozsoy2026egoexor} with corresponding gaze annotations to create EgoExOR-Gaze. EgoExOR consists of 41 takes with a total duration of 94 minutes at 15 Frames Per Second (FPS), where each take covers one or more phases of two simulated surgical procedures: ultrasound-guided needle insertion and minimally invasive spine surgery. It provides synchronized egocentric and exocentric images, 2D gaze coordinates in egocentric views, hand tracking, depth maps, point clouds, audio, and SSG annotations. These SSGs are represented as triplets containing various relations, including ``looking''. However, these annotations focus on a broad range of clinical interactions rather than specifically identifying all attended entities. Consequently, we observe that the existing ``looking'' relations are highly incomplete and unevenly distributed across the dataset. Therefore, we exhaustively annotate ``looking'' relations at 1 FPS, which captures consistent visual attention over time while keeping the annotation effort manageable. 
To ensure reliable gaze annotations, we retain only takes with all five synchronized exocentric camera views, excluding those recorded with only two views. We further exclude takes with missing egocentric gaze coordinates. 
In the end, we annotated 30 takes, including 4,432 multi-view image sets. Our gaze annotations cover four clinical roles (\emph{head surgeon}, \emph{assistant}, \emph{circulator}, and \emph{anaesthetist}) and 30 gaze target entity classes from the EgoExOR vocabulary. Since a gaze point may correspond to a region containing multiple interacting entities rather than a single isolated object, we adopt a multi-label annotation strategy to include all relevant entities. Within the 30 takes, the original EgoExOR annotations contain 2,751 ``looking'' relations. We kept 2,558 relations unchanged, corrected 173, removed 20, and added 16,457 new relations. 

During annotation, we first initialize head bounding boxes for all visible clinicians in every exocentric view using a Faster R-CNN model~\citep{ren2015faster} trained on CrowdHuman~\citep{shao2018crowdhuman} and 4D-OR~\citep{ozsoy20224d}. Using an in-house annotation tool, we manually correct missing or inaccurate detections and associate each person with their clinical role and the scene entities they are looking at. To obtain 3D gaze target annotations, we transfer 2D gaze coordinates from the egocentric images to exocentric views using RoMa v2~\citep{edstedt2026roma}, then manually verify and correct the matched points. When a gaze target is visible in at least two exocentric views, we reconstruct its 3D position through triangulation. If only one valid view is available, we back-project the gaze point using its corresponding depth value. To handle missing depth values, we complete the original EgoExOR depth maps using VGGT-$\Omega$~\citep{wang2026vggtomega}. Similarly, we reconstruct 3D head positions from the centers of the annotated 2D head bounding boxes using triangulation or back-projection.
Fig.~\ref{fig:annotations} shows an example of the annotations. Table~\ref{tab:dataset_stats} summarizes the dataset statistics and Table~\ref{tab:entity_dist} presents the distribution of gaze target entities. The EgoExOR-Gaze annotations and source code for our methods will be made publicly available.

\section{Methodology}\label{sec:method}

\begin{figure*}
\centering
\includegraphics[trim=0 10 0 10, clip, width=0.9\linewidth]{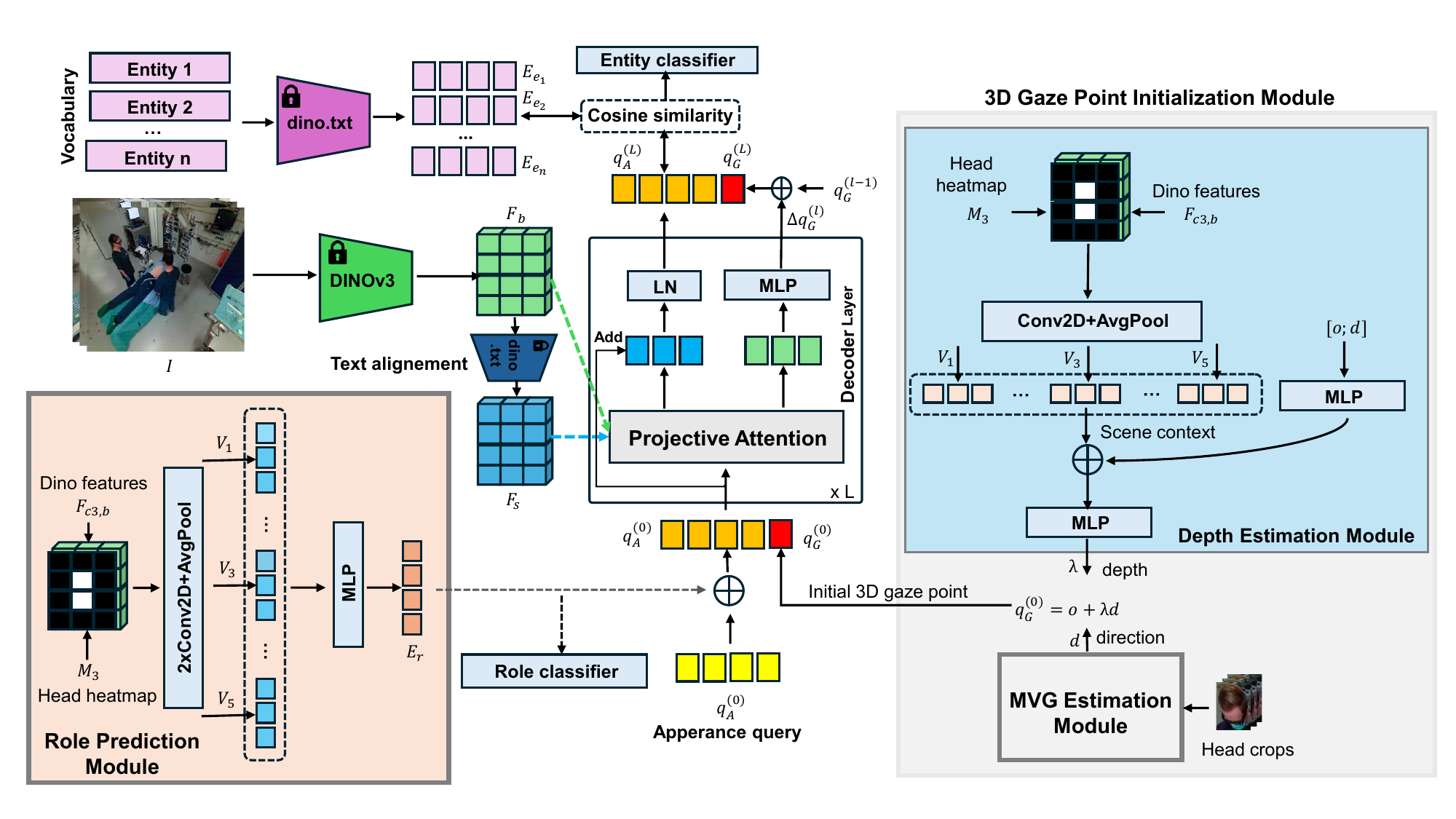}
\caption{Overview of the gaze graph generation architecture. A frozen DINOv3 backbone extracts raw and text-aligned visual features. The initial 3D gaze point is computed from the estimated gaze direction and depth. The role prediction module estimates each clinician's role, and the corresponding role embedding is added to the initial appearance query. Decoder layers iteratively refine the 3D gaze point and appearance query through projective attention. Finally, the appearance query is compared with `dino.txt' vocabulary embeddings using cosine similarity to identify attended entities.}
\label{fig:architecture}
\end{figure*}

Given calibrated multi-view images $I \in \mathbb{R}^{C \times 3 \times H \times W}$, head bounding boxes $B \in \mathbb{R}^{C \times P \times 4}$, and 3D head positions $O \in \mathbb{R}^{P \times 3}$ for $P$ persons, the goal is to predict their clinical roles $\mathcal{R}=\{r_1,\ldots,r_P\}$ and attended scene entities from $\mathcal{E}=\{e_1,\ldots,e_n\}$. Our model processes each person independently, predicting their clinical role, a 3D gaze point $q_G \in \mathbb{R}^3$, and the entities they are looking at. As illustrated in Fig.~\ref{fig:architecture}, the architecture consists of three main components: a 3D gaze point initialization module, a role prediction module, and a decoder that iteratively refines the 3D gaze point and predicts attended entities. The per-person predictions are then combined into a scene-level gaze graph, represented by a binary matrix $A_I \in \{0,1\}^{|\mathcal{R}| \times |\mathcal{E}|}$, where $A_{I,r_j,e_i}=1$ indicates that a person with role $r_j$ looks at entity $e_i$. The predicted gaze graphs are subsequently used for surgical action recognition and SSG generation.

\subsection{Gaze graph generation}

We propose a multi-view gaze graph generation model that jointly estimates each person's 3D gaze target and identifies the attended scene entities. For each camera view $c_i$, we extract raw visual features $F_{c_i,b} \in \mathbb{R}^{D \times H_f \times W_f}$ using DINOv3~\citep{simeoni2025dinov3}. We then process these features with the `dino.txt' visual head~\citep{jose2025dinov2} to obtain text-aligned features $F_{c_i,s}$. The raw and text-aligned features provide geometric and semantic information, respectively, for gaze graph generation.

\noindent \textbf{3D gaze point initialization.} We initialize the 3D gaze point by estimating the gaze direction and depth from the 3D head position. 

\noindent \textit{Multi-view gaze estimation.} Following MVGTE~\citep{miao2025multi}, we predict a 3D gaze direction and its uncertainty from each view where the head is visible, and select the direction $d \in \mathbb{R}^3$ with the lowest uncertainty.

\noindent \textit{Depth estimation.} For each view, the head heatmap $M_i$ is concatenated with the raw visual features $F_{c_i,b}$ and encoded into a view feature $V_i$ through convolution and average pooling. The concatenated multi-view features are combined with embeddings of the 3D head position $o$ and gaze direction $d$ to predict the gaze depth $\lambda$ via MLPs. The initial 3D gaze point is then computed as: 
\begin{equation}
q_G^{(0)} = o + \lambda d.
\end{equation}

\noindent \textbf{Role prediction.} Similarly, the role prediction module aggregates head-conditioned visual features across views to produce a role embedding $E_r$, which is used for clinical role classification and appearance query initialization.

\noindent \textbf{Decoder.} The decoder iteratively refines a set of two components, $q^{(l)} = \{q_A^{(l)}\in \mathbb{R}^{D_q}, q_G^{(l)}\}$, consisting of an appearance query and a 3D gaze point, over $L$ layers. The appearance query is initialized with a learned embedding, optionally augmented with the predicted role embedding $E_r$. Each decoder layer uses projective attention~\citep{zhang2021direct} to sample local visual features around the projection of the current 3D gaze point. Specifically, $q_G^{(l-1)}$ is projected into camera $c_i$ to obtain an anchor $p_{c_i}^{(l)}$, around which $K$ features are sampled and aggregated as: 
\begin{equation}
f_{c_i} = \sum_{k=1}^{K} \alpha_k \cdot \big[F_{c_i}(p_{c_i}^{(l)} + \Delta p_k) + f_{\text{anc}, c_i}\big].
\label{eq}
\end{equation}
Here, the sampling offsets $\Delta p_k$ and attention weights $\alpha_k$ are predicted from the appearance query and anchor feature $f_{\text{anc},c_i}$. The sampled features are then aggregated across views using learned weights.

\noindent \textit{Appearance query update.} We apply projective attention to the text-aligned features $F_{c_i,s}$ and update $q_A$ through a residual linear projection followed by layer normalization. At the final layer, we compare the refined appearance query with `dino.txt' vocabulary embeddings~\citep{jose2025dinov2} using cosine similarity, then apply a sigmoid activation to predict attended entities. Class-specific thresholds are selected to maximize the validation F1 scores. 

\noindent \textit{Geometry query update.} In parallel, projective attention is applied to the raw features $F_{c_i,b}$. To incorporate person-specific scene context, we add a learned head embedding at the visible head locations and process the resulting multi-view feature maps through convolutional layers. The scene context and sampled local features are combined to predict a 3D offset $\Delta q_G^{(l)}$, updating the gaze point as: 
\begin{equation}
q_G^{(l)} = q_G^{(l-1)} + \Delta q_G^{(l)}.
\label{eq:3d_update}
\end{equation}

\noindent \textbf{Training objective.} We use cosine loss for gaze direction, $L_1$ losses for depth and 3D gaze position, cross-entropy for role classification, and asymmetric loss~\citep{ridnik2021asymmetric} for entity prediction. The final loss function is:
\begin{equation}
\mathcal{L} = \lambda_{\text{dir}}\mathcal{L}_{\text{dir}} + \lambda_{\text{depth}}\mathcal{L}_{\text{depth}} + \lambda_{\text{pos}}\mathcal{L}_{\text{pos}} + \lambda_{\text{role}}\mathcal{L}_{\text{role}} + \lambda_{\text{entity}}\mathcal{L}_{\text{entity}}.
\label{eq:final_loss}
\end{equation}

\subsection{Downstream tasks}

We evaluate predicted gaze graphs on two downstream tasks: fine-grained surgical action recognition and SSG generation. 

\noindent \textbf{Surgical action recognition.} Given a sequence of $T$ frames, we represent the gaze graphs as a binary tensor $A \in {0,1}^{T \times |\mathcal{R}| \times |\mathcal{E}|}$, where $A_{t,r,e}$ indicates whether role $r$ looks at entity $e$ at time $t$. We assign learnable embeddings $v_e \in \mathbb{R}^d$ and $v_r \in \mathbb{R}^d$ to each entity and role, respectively. The gaze feature for role $r$ at time $t$ is computed as:
\begin{equation}
h_{t,r} = v_r + \frac{\sum_{e \in \mathcal{E}} A_{t,r,e} v_e}{\max\left(1, \sum_{e \in \mathcal{E}} A_{t,r,e}\right)}.
\end{equation}
A spatial transformer encoder models interactions between roles at each frame, followed by a temporal Transformer encoder that processes each role's sequence independently. Following the action taxonomy of OR-Action~\citep{tristram2026or}, we consider $J$ classes, including idle and $J-1$ surgical actions. Two MLP classifiers predict the activity status (active or idle) and the surgical action, respectively. Both are trained using cross-entropy losses, with the total loss defined as: 
\begin{equation}
    \mathcal{L}_{\text{action-rec}} = \lambda_{\text{activity}}\mathcal{L}_{\text{activity}} + \lambda_{\text{action}}\mathcal{L}_{\text{action}}.
    \label{eq:action_loss}
\end{equation}

\noindent \textbf{Semantic scene graph generation.} We adapt the EgoExOR architecture~\citep{ozsoy2026egoexor} by replacing its egocentric branch with a gaze graph branch while keeping the exocentric branch unchanged. At each frame, each predicted ``looking'' relation $(r,\text{looking},e)$ is encoded as a gaze token by summing the learnable embeddings of its role and entity. These tokens replace the egocentric branch tokens and are combined with the exocentric features for SSG generation. 

\section{Experiments}\label{sec:experiments}

\noindent \textbf{Datasets.} We conduct experiments on EgoExOR-Gaze, consisting of 30 takes with gaze graph annotations and surgical action labels from OR-Action~\citep{tristram2026or}. Following EgoExOR, we use 19, 6, and 5 takes for training, validation, and testing, respectively. For gaze graph generation and SSG generation, we train and evaluate the models at 1~FPS. For surgical action recognition, we apply two settings: 1~FPS to align with our gaze annotations and 4~FPS to match the temporal sampling rate used in the original OR-Action experiments. For SSG generation, we exclude ``looking'' relations from both training targets and evaluation to focus on predicting other semantic relations.

\noindent \textbf{Evaluation metrics.} We report macro-F1 for entity prediction, role prediction, gaze graph generation, and SSG generation. For gaze graph generation, we additionally report Exact Match Ratio (EMR), which considers a prediction correct only if the complete set of gaze target entities for a given clinician exactly matches the ground truth. For surgical action recognition, we report frame accuracy excluding the idle class, Edit Score, and segmental F1 scores at overlap thresholds of 0.1, 0.25, and 0.5.

\noindent \textbf{Implementation details.} For gaze graph generation, we use a frozen DINOv3 ViT-L/16 backbone~\citep{simeoni2025dinov3} and dino.txt~\citep{jose2025dinov2} with $H = W = 336$, $H_f=W_f=21$ and $D = D_q = 1024$. The decoder consists of $L = 4$ layers and uses $K = 5$. We train the model for 40 epochs using AdamW with a batch size of 8. The learning rate increases linearly from $10^{-5}$ to $10^{-4}$ over three warmup epochs and then follows a cosine decay to $10^{-6}$. We set $\lambda_{\mathrm{dir}}$, $\lambda_{\mathrm{depth}}$, $\lambda_{\mathrm{pos}}$, $\lambda_{\mathrm{role}}$, $\lambda_{\mathrm{entity}}$ to 0.2, 0.5, 2.5, 0.1, and 1.5, respectively, in Eq.~\ref{eq:final_loss}. For surgical action recognition, we use a 16-second temporal window for both 1~FPS and 4~FPS settings, with an embedding dimension $d = 256$. We train the model for 60 epochs using AdamW with a batch size of 6 and cosine learning-rate decay from $10^{-4}$ to $10^{-6}$. Both loss weights $\lambda_{\mathrm{activity}}$ and $\lambda_{\mathrm{action}}$ are set to 1 in Eq.~\ref{eq:action_loss}. $J$ is 78.

\subsection{Results}
\noindent \textbf{Gaze graph generation.} We compare our model with two variants of EgoExOR: (1) using only the exocentric branch and (2) using both exocentric and egocentric branches, with additional egocentric images and eye-tracking gaze coordinates as inputs. We also report an oracle variant by using the ground-truth 3D gaze point as input to the projective decoder. As shown in Tab.~\ref{tab:gg_results}, our model, which relies solely on exocentric images, outperforms both baselines. EgoExOR generates scene graph triplets from globally aggregated features, while our approach explicitly samples local visual features across multiple views based on the estimated 3D gaze point. These results indicate that geometry-guided local feature sampling is more effective for identifying looked-at entities than global triplet generation. 

\begin{table}[t]
\centering
\caption{Comparison of gaze graph generation performance.}
\label{tab:gg_results}
\resizebox{0.4\textwidth}{!}{
\begin{tabular}{lcc}
\toprule
Model & EMR & Macro-F1 \\
\midrule
Ours (exo images + 3D gaze GT) & 71.14 & 72.01 \\
\midrule
EgoExOR (exo branch)~\citep{ozsoy2026egoexor} & 8.89 & 15.02 \\
EgoExOR (exo + ego branches)~\citep{ozsoy2026egoexor} & 39.30 & 41.51 \\
Ours (exo images) & \textbf{60.14} & \textbf{52.65} \\
\bottomrule
\end{tabular}
}
\end{table}

\noindent \textbf{Surgical action recognition.} We compare our model against OR-Action~\citep{tristram2026or} using either exocentric or egocentric video inputs. As shown in Table~\ref{tab:action_results}, our model using predicted gaze graphs outperforms both baselines across all metrics at both 1~FPS and 4~FPS, with particularly strong improvements in temporal metrics. These results demonstrate that gaze-based relations provide valuable contextual information for recognizing surgical actions and producing more temporally coherent action predictions. We also report an oracle variant using ground-truth gaze graphs at 1~FPS to assess the impact of gaze graph prediction errors. %Fig... illustrates the temporal consistency of our model.

\begin{table}[t]
\centering
\caption{Surgical action recognition results at 1~FPS and 4~FPS on the test set (GG = Gaze Graph).}
\label{tab:action_results}
\resizebox{0.48\textwidth}{!}{
\begin{tabular}{l|l|ccccc}
\toprule
& Model & Acc. & Edit & F1@0.1 & F1@0.25 & F1@0.5 \\
\midrule
\parbox[c]{7mm}{\multirow{4}{*}{\makecell{1FPS}}} & Oracle (Ours + GT GG) & 69.42 & 65.16 & 53.91 & 50.37 & 36.95 \\
\cmidrule{2-7}
& OR-Action (Exo)~\citep{tristram2026or} & 54.50 & 27.06 & 30.21 & 24.30 & 12.81 \\
& OR-Action (Ego)~\citep{tristram2026or} & 66.12 & 35.36 & 32.36 & 28.73 & 19.11 \\
& Ours (predicted GG) & \textbf{66.55} & \textbf{57.81} & \textbf{43.44} & \textbf{41.68} & \textbf{28.49} \\
\midrule
\parbox[c]{7mm}{\multirow{3}{*}{\makecell{4FPS}}} & OR-Action (Exo)~\citep{tristram2026or} & 51.86 & 25.77 & 30.38 & 24.53 & 14.52 \\
& OR-Action (Ego)~\citep{tristram2026or} & 69.95 & 34.75 & 35.39 & 31.73 & 24.28 \\
& Ours (predicted GG) & \textbf{70.69} & \textbf{57.31} & \textbf{38.46} & \textbf{33.86} & \textbf{25.36} \\
\bottomrule
\end{tabular}
}
\end{table}

\begin{table*}[t]
\caption{Per-predicate F1 scores for SSG generation (GG = Gaze Graph). \textbf{Best scores}, \underline{second best scores}.}
\label{tab:ssg_results}
\centering
\resizebox{0.9\textwidth}{!}{
\begin{tabular}{@{}l|*{16}{c}|c@{}}
\toprule
\rotatebox{90}{\textbf{Relation}}
& \rotatebox{90}{anaesthetising}
& \rotatebox{90}{applying}
& \rotatebox{90}{aspirating}
& \rotatebox{90}{closeto}
& \rotatebox{90}{controlling}
& \rotatebox{90}{cutting}
& \rotatebox{90}{disinfection}
& \rotatebox{90}{entering}
& \rotatebox{90}{holding}
& \rotatebox{90}{injecting}
& \rotatebox{90}{inserting}
& \rotatebox{90}{lyingon}
& \rotatebox{90}{manipulating}
& \rotatebox{90}{removing}
& \rotatebox{90}{scanning}
& \rotatebox{90}{touching}
& \rotatebox{90}{\textbf{Macro-F1}} \\
\midrule
Exo only
& \underline{0.11} & 0.48 & 0.11 & \underline{0.95}
& \textbf{0.93} & 0.00 & 0.19 & 0.23
& \underline{0.47} & 0.00 & 0.17 & \textbf{1.00}
& 0.26 & 0.00 & \textbf{0.98} & \underline{0.82}
& 0.42 \\

Exo+\textbf{GG}
& \textbf{0.19} & \underline{0.61} & \underline{0.66} & \textbf{0.96}
& 0.89 & \underline{0.21} & \underline{0.54} & \underline{0.26}
& 0.45 & 0.00 & \underline{0.34} & \textbf{1.00}
& \textbf{0.44} & \underline{0.04} & \textbf{0.98} & 0.80
& \underline{0.52} \\

EgoExOR
& 0.00 & \textbf{0.86} & \textbf{0.83} & \textbf{0.96}
& \underline{0.91} & \textbf{0.30} & \textbf{0.65} & \textbf{0.61}
& \textbf{0.77} & \textbf{0.88} & \textbf{0.77} & \underline{0.99}
& \underline{0.39} & \textbf{0.34} & \textbf{0.98} & \textbf{0.93}
& \textbf{0.70} \\
\bottomrule
\end{tabular}
}
\end{table*}

\noindent \textbf{Semantic scene graph generation.} We compare three input configurations: exocentric views only, exocentric views with predicted gaze graphs, and the full EgoExOR model with both exocentric and egocentric inputs. As shown in Table~\ref{tab:ssg_results}, incorporating gaze graphs as input improves the macro-F1 score from 0.42 to 0.52, though it remains below the full EgoExOR model. Specifically,
gaze graphs improve the recognition of relations such as \emph{aspirating}, \emph{disinfection}, and \emph{cutting}, where scene-level visual attention provides useful cues. However, they are less effective for relations such as \emph{removing} and \emph{injecting}, where different actions may involve attention to the same entities and require finer visual details to distinguish. These results suggest that gaze graphs offer a practical alternative for improving SSG generation when egocentric views are unavailable.

\subsection{Ablation Study}

We evaluate three components of our gaze graph generation model in Table~\ref{tab:ablation}.

\noindent \textbf{Role conditioning.} Role-conditioned appearance queries improve both metrics, suggesting that predicted roles help identify attended entities.

\noindent \textbf{Random depth.} Replacing the estimated depth of 3D gaze point with a random value sampled from the training set distribution reduces performance, highlighting the importance of depth estimation for 3D gaze point initialization.

\noindent \textbf{Text alignment.} Replacing the text-aligned features $F_{c_i,s}$ with raw DINOv3 features $F_{c_i,b}$ obtains comparable macro-F1 but lower EMR, indicating that text alignment helps predict the complete set of attended entities.

\begin{table}[t]
\centering
\caption{Ablation study of our gaze graph generation architecture.}
\label{tab:ablation}
\resizebox{0.48\textwidth}{!}{
\begin{tabular}{l|cc}
\toprule
Model & EMR & Macro-F1 \\
\midrule
w/o role conditioning & 58.45 & 48.88 \\
using random depth & 53.99 & 49.30 \\
using raw DINOv3 features w/o text alignment & 51.69 & 52.46 \\
\midrule
Full model & \textbf{60.14} & \textbf{52.65} \\
\bottomrule
\end{tabular}
}
\end{table}

\section{Conclusion}\label{sec:conclusion}

In this paper, we investigate gaze graphs as a scene-level representation of visual attention that connects clinicians to attended scene entities through ``looking'' relations. We extend EgoExOR with gaze annotations to create EgoExOR-Gaze, the first benchmark for multi-view gaze graph generation in the OR, and propose a novel architecture using only exocentric views. Experimental results demonstrate that predicted gaze graphs improve fine-grained surgical action recognition and semantic scene graph generation, highlighting the importance of explicitly modeling visual attention in surgical scenes even when egocentric views are unavailable.

\section{Acknowledgment}\label{sec:acknowledgment}

This work was supported by French state funds managed within the Plan Investissements d’Avenir by the ANR under references ANR-22-FAI1-0001 (project DAIOR), ANR-23-IACL-0004 (ENACT AI Cluster), ANR-10-IAHU-02 (IHU Strasbourg) and by BPI France (Project 5G-OR). 
This work was granted access to the servers/HPC resources managed by CAMMA, IHU Strasbourg, Unistra Mesocentre, and GENCI-IDRIS [AD011014722R2, AD011011631R4, and AD011011638R3].

\bibliographystyle{model2-names.bst}
\bibliography{sn-bibliography}

\end{document}